\documentclass[10pt,twocolumn,letterpaper]{article}

\usepackage{iccv}
\usepackage{times}
\usepackage{epsfig}
\usepackage{graphicx}
\usepackage{amsmath}
\usepackage{arydshln }
\usepackage{amssymb}
\usepackage{enumerate}
\usepackage{multirow}
\usepackage{times}
\usepackage{epsfig}
\usepackage{graphicx}
\usepackage{amsmath}
\usepackage{amssymb}
\usepackage{graphicx}
\usepackage{amsmath}
\usepackage{mathrsfs}
\usepackage{booktabs}
\usepackage{gensymb}
\usepackage{esvect}
\usepackage{algorithm}
\usepackage{bm}
\usepackage{subfigure}
\usepackage{pifont}

\usepackage{algpseudocode}

\usepackage[pagebackref=true,breaklinks=true,letterpaper=true,colorlinks,bookmarks=false]{hyperref}

 \iccvfinalcopy 

\def\iccvPaperID{1587} 

\ificcvfinal\fi

\begin{document}

\newcommand{\yes}{\ding{51}}
\newcommand{\no}{\ding{55}}

\newcommand{\todo}[1]{\textcolor{red}{TODO: #1}}
\newcommand{\TODO}[1]{\textcolor{red}{\textbf{****** #1 ******}}}
\newcommand{\hongmin}[1]{{\color{cyan}\textsc{Hongmin:} #1}}
\newcommand{\jianchao}[1]{{\color{blue}\textsc{Jianchao:} #1}}
\newcommand{\jiliu}[1]{{\color{blue}\textsc{Jiliu:} #1}}
\newcommand{\zhangxiong}[1]{{\color{green}\textsc{XiongZhang:} #1}}

\title{Hand Image Understanding via Deep Multi-Task Learning}

\author{Xiong Zhang$^{1 }$\thanks{corresponding author, zhangxiong@yy.com}, Hongsheng Huang$^{2}$, Jianchao Tan$^3$, Hongmin Xu$^{4}$,\\
 Cheng Yang$^1$, Guozhu Peng$^1$, Lei Wang$^1$, Ji Liu$^3$\\
\normalsize{$^1$YY Live, Baidu Inc., $^2$Joyy Inc., $^3$AI Platform, Kwai Inc., $^4$OPPO Inc.,}\\
}

\maketitle

\begin{abstract}
Analyzing and understanding hand information from multimedia materials like images or videos is important for many real world applications and remains active in research community. There are various works focusing on recovering hand information from single image, however, they usually solve a single task, for example, hand mask segmentation, 2D/3D hand pose estimation, or hand mesh reconstruction and perform not well in challenging scenarios. To further improve the performance of these tasks, we propose a novel Hand Image Understanding (\textbf{HIU}) framework to extract comprehensive information of the hand object from a single RGB image, by jointly considering the relationships between these tasks. To achieve this goal, a cascaded multi-task learning (MTL) backbone is designed to estimate the 2D heat maps, to learn the segmentation mask, and to generate the intermediate 3D information encoding, followed by a coarse-to-fine learning paradigm and a self-supervised learning strategy.
Qualitative experiments demonstrate that our approach is capable of recovering reasonable mesh representations even in challenging situations.
Quantitatively, our method significantly outperforms the state-of-the-art approaches on various widely-used datasets, in terms of diverse evaluation metrics.
\end{abstract}

\section{Introduction}
Hand image understanding (HIU) keeps active in both computer vision and graphics communities, aiming to recover the spatial configurations from RGB/depth images, including 2D/3D hand pose estimation, hand mask segmentation  and hand mesh reconstruction, 
which have been employed in various domains \cite{sridhar2015investigating,markussen2014vulture,hurst2013gesture,jang20153d,liang1998real,starner1998real}.
Recovering the spatial configurations is still challenging, due to the inherent depth and scale ambiguity, diverse appearance variation, and complicated articulations.
While a bunch of existing works have considered markerless HIU, most of whom require depth camera \cite{oikonomidis2011efficient,xu2013efficient,sridhar2013interactive,qian2014realtime,khamis2015learning,ge2016robust,yuan2018depth,li2019point,malik2020handvoxnet,fangjgr,huang2020hot,huanghand} or synchronized multi-view images \cite{ballan2012motion,gomez2019large,simon2017hand,sridhar2013interactive,wang20116d} to deal with the aforementioned challenges.
As a consequence, most of those methods are impractical in real-world situations where only monocular RGB images are available.
\begin{figure}
\centering
\includegraphics[width=0.4\textwidth]{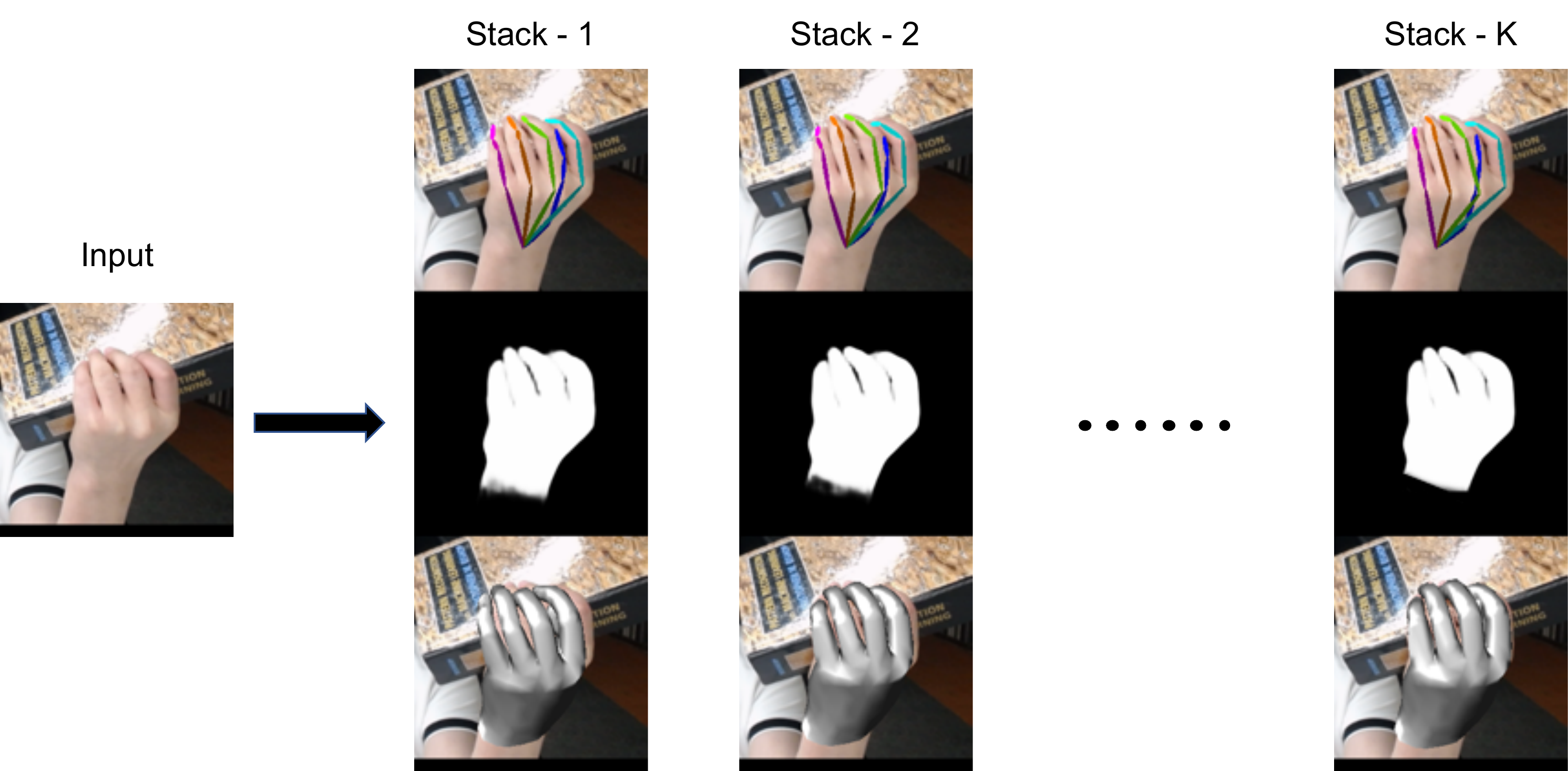}
\caption{{\bf Hand Image Understanding.}
The figure illustrates the fundamental idea of this work. We derive the 2D hand pose, hand mask, hand mesh (including 3D hand pose) representation simultaneous from a monocular RGB image of the hand object in a coarse-to-fine manner.
}
\label{fig:illustrator}
\end{figure}

For the monocular RGB scenario, the main obstacles reside in three-folds.
Firstly, the lack of high-quality large-scale datasets with precise annotations.
Existing datasets are either synthesized using software \cite{zimmermann2017learning,hasson2019learning,mueller2018ganerated}, or labeled in a semi-automated manner \cite{zimmermann2019freihand,kulon2020weakly}, 
or collected in a controlled experimental environment \cite{yu2020humbi,zhang2017hand,sridhar2016real}.
Secondly, the incapability of the current datasets makes the trained models not generalize well to various wild images, especially under self occlusions and complex configurations, which may hinder its applications.
Thirdly, contemporary approaches fail to exploit unlabeled images, which is more widely distributed than those with annotations.

The above obstacles motivate us to bring out two questions: 
\emph{Can the existing multi-modality data be harnessed comprehensively to tackle the aforementioned difficulties? }
\emph{Can the tremendous wild images without labels be exploited well enough to favor the HIU?}

In this work, we demonstrate the answers to be \textbf{yes}, and the fundamental idea is illustrated in Figure \ref{fig:illustrator}.
Specifically, an innovative multi-task learning (MTL) framework is designed to tackle the HIU problem, which follows the cascade coarse-to-fine design manner.
Concretely, the framework consists of a backbone and several regressor heads corresponding to different tasks.
The backbone aims to learn the various elementary representations from hand images, including 2D pose estimation, hand mask segmentation, and 3D part orientation field (POF) encoding.
To reconstruct the whole hand mesh efficiently, we exploit the generative hand model MANO \cite{romero2017embodied}, and employ the regressor heads to regress MANO's parameters based on the semantic features of the elementary tasks.
To efficiently fuse beneficial semantic-features among various tasks, we conceive the task attention module (TAM) to aggregate the semantic features across individual tasks and derive compact high-level representations by removing redundant features.
Note that the 3D hand joints can be achieved as a side output in MANO. 
With these designs together, one can obtain the 2D/3D hand pose, hand mask, and hand mesh from a RGB image simultaneously.

It is clear that the whole framework can be trained with generic supervised learning by leveraging existing multi-modality datasets.
The self-supervised learning strategies can be adopted by leveraging the implicit relationship constraints maintained among the reasonable predictions from each task.
For instance, the mask rendered from the hand mesh with proper camera parameters, shall match the one that is estimated by the backbone; the coordinates of the re-projected 2D hand pose shall be close to the integral of locations encoded in the heat-maps.
The self-supervised learning make it possible to exploit enormous wild images, which can improve the accuracy of the framework, and enhance the generalization capability.
Additionally, considering the absence of a large-scale hand dataset with well-labeled  hand mask and 2D pose, we collect a high-quality dataset with manually labeled 2D hand pose and hand mask.

To summarize, our main contributions are as bellows:
\begin{enumerate}[$\bullet$]
\item We design an innovative cascade multi task learning (MTL) framework, dubbed HIU-DMTL, for hand image understanding, which can exploit existing multi-modality hand datasets effectively.
  \item The self-supervised learning (SSL) was firstly introduced to alleviate the insufficient data problem for HIU, and the effectiveness of which has been verified through extensive experiments comprehensively.
  
  \item We propose a simple while effective task attention module (TAM), targeting to aggregate semantic features across various tasks, which proves to be instrumental for MTL on the HIU tasks.

  \item Our HIU-DMTL framework outperforms contemporary hand mesh recovery approaches \cite{hasson2019learning,ge20193d,boukhayma20193d,zhang2019end,zhou2020monocular,moon2020i2l}, 
  and demonstrates new state-of-the-art performances on 
  widely used benchmarks \cite{zhang2017hand,zimmermann2017learning,sridhar2016real,zimmermann2019freihand,simon2017hand}, in terms of various evaluation metrics.
      
\end{enumerate}

\begin{figure*}
\centering
\includegraphics[width=0.95\textwidth]{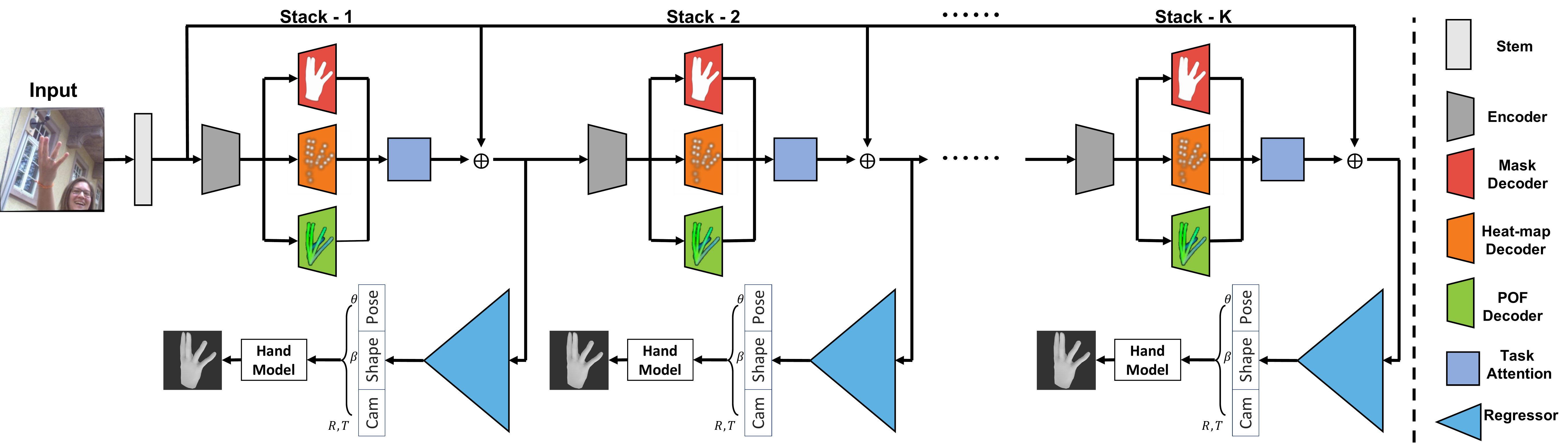}
\caption{{\bf Framework Architecture.}
\label{fig:mainframe}
The whole pipeline follows the classical cascade coarse-to-fine design paradigm, which consists of two main components:
(1) a novel multi-task learning backbone aiming to estimate heat-maps that encode the 2D hand pose, to learn the hand segmentation mask, and to generate the POF encoding that covers 3D information,
(2) the regressor heads aiming to regress 3D parameters $\Theta=\{\theta, \beta, R, T\}$, based on the multi-task features, of the parametric hand model MANO \cite{romero2017embodied} and the perspective camera.
}
\end{figure*}

\section{Related Work}
Due to the extensive scope of related works, it is hard to summarize all of them comprehensively. We only discuss works that are strongly related with our framework settings.

\textbf{3D Hand Pose Estimation.}
The pioneering work \cite{zimmermann2017learning} firstly applies the deep learning technique to estimate the 3D hand pose from a single RGB image.
Since then, 3D hand pose estimation draws a great deal of attention from the community \cite{panteleris2018using,mueller2018ganerated,cai2018weakly,iqbal2018hand,spurr2018cross,yang2019disentangling,cai2019exploiting,theodoridis2020cross,zhao2020knowledge,spurr2020weakly,fan2020adaptive}.
Those methods either target to solve the perspective ambiguity problem by introducing geometry-constraints of hand articulation
\cite{zimmermann2017learning,panteleris2018using,mueller2018ganerated,spurr2020weakly}, 
to investigate intricate learning strategies to achieve better performance \cite{spurr2018cross,iqbal2018hand,cai2018weakly,yang2019disentangling,cai2019exploiting,theodoridis2020cross,fan2020adaptive},
or to tackle the challenge of lacking sufficient high-quality hand data \cite{zimmermann2017learning,mueller2018ganerated,yang2019disentangling,zhao2020knowledge}.
Despite the significant progress have been achieved, estimating 3D hand pose from a monocular RGB image remains to be challenging, and the lacking of sufficient well-labeled data is still one of the main obstacles.

\textbf{Hand Mesh Recovery.}
Besides the hand pose estimation, hand mesh recovery is another essential and active research topic.
One line of works focus on reconstructing the hand mesh representation in a general situation.
For instance, some works \cite{boukhayma20193d,baek2019pushing,zhang2019end,zimmermann2019freihand,kulon2020weakly,zhou2020monocular,yang2020seqhand} capture the hand mesh representation by regressing the generative hand model MANO's parameters, while others \cite{ge20193d,moon2020i2l} instead turn to estimate the 3D mesh vertices directly in order to learn nonlinear hand shape variations.
Another line of works \cite{hasson2019learning,baek2020weakly,huang2020hot} attempt to reconstruct the hand mesh in the hand-object interaction environment, in which the hand and object are reconstructed jointly by introducing interactive relationship constraints. 
Though the above approaches can recover reasonable hand mesh in experimental benchmarks, in practice, we find that approaches \cite{boukhayma20193d,hasson2019learning,moon2020i2l,ge20193d,zhang2019end,zhou2020monocular}, with the publicly released code and pre-trained model, do not work well across different datasets or generalize well to the real-world situations.

\textbf{Multi-Task Learning.}
Multi-task learning (MTL) is one methodology to improve the performance of tasks 
by utilizing the limited training samples and sharing the beneficial information 
among all tasks, which has been successfully adopted in many domains \cite{collobert2008unified,girshick2015fast,russakovsky2015imagenet}.
One broadly employed MTL method is hard-parameter sharing, which trains one shared encoder, followed by multiple task-specific decoders for different tasks. 
Some of them also further design a decoder-fusing module to distill information of different tasks to refine the final tasks' prediction.
Recent works  \cite{rogez2017lcr, popa2017deep,lin2017recurrent, tome2017lifting,wan2018dense, du2019crossinfonet, chen2020nonparametric} have applied such multi-task framework into pose estimation tasks and achieved state-of-the-art performance.
Our approach follows general settings of previous multi-task learning methods, which makes use of an encoder, several certain task-specific decoders, and a features fusing module. 
Solving HIU tasks under cascade multi-stack MTL framework still remains less explored in community, our work demonstrates such combination can achieve SOTA performance. 



\section{Framework}
The primary goal of this work is to design a unified framework to provide comprehensive information of the hand object from a single RGB image, 
including 2D hand pose, 3D hand joints, hand mask, and hand mesh representation.
To achieve this goal, a multi-stacked multi-branch backbone is designed to learn various elementary representations of hand object, then the mesh reconstruction is achieved by estimating parameters of the parametric model MANO \cite{romero2017embodied} based on the elementary representations. 
The whole framework is illustrated in Figure \ref{fig:mainframe}.

\subsection{Backbone}
The backbone consists of a stem module and several MTL blocks sharing the same structure.
Each MTL block consists of an encoder that is shared by all task-specified decoders in that block, several certain dedicated decoders that aim to tackle individual primary task, and a TAM that aggregating features across various tasks.
In practice, we take the 2D hand pose, the hand mask, and the POF encoding as the learning target of the intermediate elementary tasks.

\textbf{Stem Module.} 
The stem module aims to extract low-level semantic features that are shared by succeeding modules.
To keep the stem module as simple as possible while be capable of covering sufficient information, 
we implement the stem module with two $7\times7$ convolution layers with stride 2, which may quickly downsample the feature-maps to reduce the computation while obtaining wide enough receptive field.

\textbf{Encoder.}
The encoder targets at generating high-level features to support various individual tasks, which takes the low-level features from the stem module and the aggregated high-level representations from the preceding TAM together as inputs.
For the first MTL block (denoted as stack $1$ in Figure \ref{fig:mainframe}), since no preceding MTL block exists, the high-level semantic features remain $\bm{0}$ with a proper size. 

\textbf{Heat-Map Decoder.}
The purpose of heat-map decoder branch is to perform 2D hand pose estimation.
Similar to recent methods \cite{cai2018weakly,ge20193d,boukhayma20193d,zhang2019end,zhou2020monocular,moon2020i2l},
we employ the 2D Gaussian-like heat-maps $\textbf{H} \in \mathbb{R}^{K \times H\times W}$ to encode the 2D hand pose, where $K$ indicates the number of joints and $\{H, W\}$ are the resolution size.
Each key point corresponds to a heat-map. The pixel value at $(x, y)$ is defined as $\exp\{-\frac{(x-\widehat{x})^2+(y-\widehat{y})^2}{2\sigma^2}\}$, where the $(\widehat{x}, \widehat{y})$ refers to the ground-truth location, corresponds to the confidence score of the key point locating in this 2D position $(x,y)$. With this heat-map, one may derive the $k_{th}$ key point with $\mathop{\arg\max}_{\{h,w\}} \textbf{H}_{(k,h,w)}$, or in a differentiable form,
\begin{align}
\sum_{h=1}^{H}\sum_{w=1}^{W}\textbf{H}_{(k,h,w)}{(h,w)}/\sum_{h=1}^{H}\sum_{w=1}^{W}\textbf{H}_{(k,h,w)}.
\label{eq:dif_pose}
\end{align}

\textbf{ Mask Decoder.}
The hand mask branch proves to be indispensable in HIU tasks, because the segmentation mask may further boost the performance of key point detection and vice versa \cite{tripathi2017pose2instance,zhang2019pose2seg,he2017mask}.
More importantly, the hand mesh may deform to best fit joint locations and may ignore hand geometrical properties, leading to an unreasonable hand mesh representation when imposing supervision over 2D/3D hand pose only  \cite{boukhayma20193d,zhang2019end,zhou2020monocular}. 
Fortunately, the mask branch not only exploits samples that are labeled with masks which lead to better performance, but also refines the hand mesh and camera parameters by penalizing the misalignment errors between the rendered mask and the estimated segmentation mask via self-supervised learning.

\textbf{POF Decoder.}
To bridge the gap between the 2D semantic features and the 3D information, we introduce the part orientation field (POF) to encode the 3D orientation of the articulated structure in 2D image space.
In practice, we introduce the standard hand skeleton hierarchy $\mathbb{S}$ data structure, which consists of a set of `(parent, child)' pairs.
For a specific bone $(A,B) \in \mathbb{S}$, where $A$ and $B$ are two joints, 
and denote $\overrightarrow{AB_{3d}}$ and  $\overrightarrow{AB_{2d}}$ as the normalized orientation from joint A to joint B in 3D and 2D domain respectively.
Then, for the bone $(A,B)$, its POF, represented as \textbf{L}$_{(a,b)} \in \mathbb{R}^{3\times H\times W}$, encodes the 3D semantic information as a 3-channel feature-map,
and the value of  \textbf{L}$_{a,b}$ at location x is defined as,
\begin{align}
\textbf{L}_{(a,b)}(\text{x})=\begin{cases}
  (\overrightarrow{AB_{2d}}, ~~ \overrightarrow{AB_{3d}}.z)& \text{x} \in \text{bone}\\
  0& \text{otherwise.}
\end{cases}
\end{align}
We shall point out that, POF values are non-zero only for the pixels belonging to the current target bone part, 
and we employ a different but more appropriate definition, comparing with \cite{xiang2019monocular}, because our POF encoding can exploit numerous wild training samples with only 2D labels.

\begin{figure}
\centering
\includegraphics[width=0.4\textwidth]{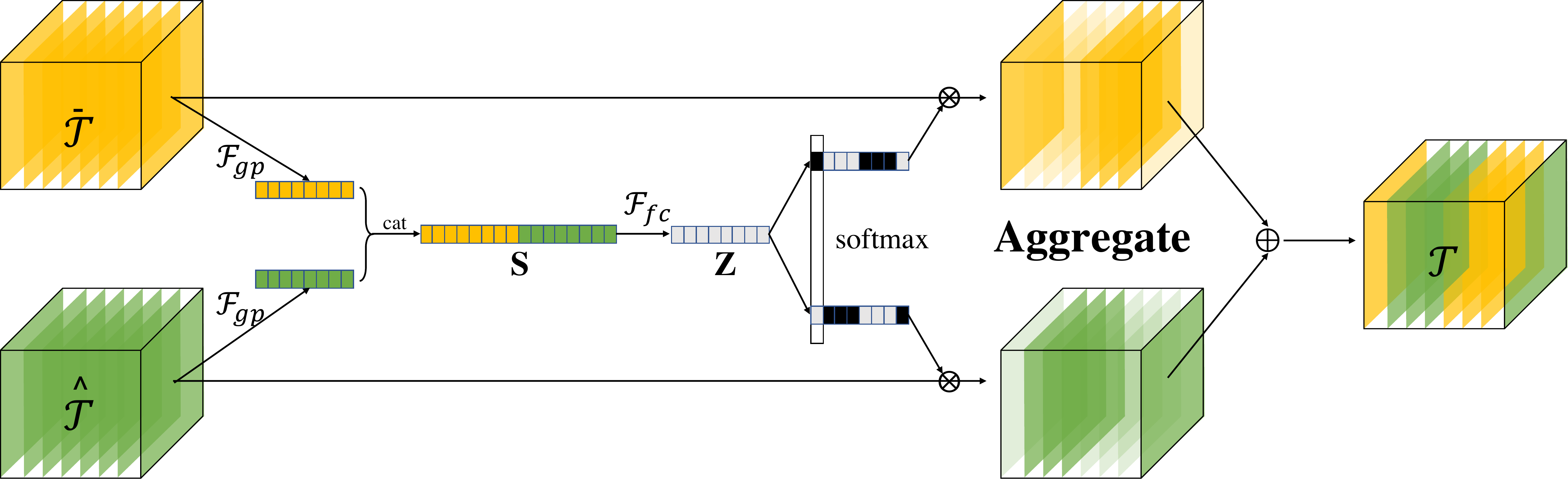}
\caption{{\bf Task Attention Module.}
\label{fig:mask_attention}
The figure illustrates the data flow of the task attention module, where $\mathcal{F}_{gp}$ and $\mathcal{F}_{fc}$ represents the global average pooling and the fully connected layers.
}
\end{figure}
\textbf{Task Attention Module.}
The TAM aims to bring together semantic features across individual tasks,
which can be formalized as a transformation, $\mathbb{R}^{N\times C \times H \times W}\mapsto \mathbb{R}^{C \times H \times W}$, where $N$ refers to the number of tasks, and $\{C, H, W\}$ denotes the spatial resolution of the feature-maps. Figure \ref{fig:mask_attention} demonstrates the structure of TAM (with $N=2$ for simplicity).
Frankly speaking, our design is motivated by SKNet \cite{li2019selective}, but with several meaningful and reasonable modifications from original setting. The beginning element-wise addition step is replaced by a global average pooling and a feature concatenation. 
Such modest but necessary adjustments make the TAM more suitable to select critical semantic features among various tasks at the expense of additional but negligible computation burden.

\subsection{Regressor Heads}
The goal of the regressor heads is to reconstruct the hand mesh surfaces.
To achieve this goal, we exploit the generative hand model MANO\cite{romero2017embodied}, and estimate the parameters of MANO that governs the mesh representation.

The mesh surface of MANO can be fully deformed and posed by the shape parameters $\beta \in \mathbb{R}^{10}$ and pose parameters $\theta \in \mathbb{R}^{15 \times 3}$, where $\beta$ models hand shape and $\theta$ represents joints rotation.
Given a pair of parameter $\{\beta, \theta\}$, the shape deformation $B_S(\beta): \mathbb{R}^{10} \mapsto \mathbb{R}^{N \times 3}$ outputs the blended shape to characterize the identity subject and pose deformation $B_P{(\theta)} : \mathbb{R}^{15 \times3 } \mapsto \mathbb{R}^{N \times 3}$ is applied to the mean template $\rm \bm \bar{T}$. Then we can obtain final mesh by rotating each bone part around joints $J(\beta)$ with the standard blend skinning function $W(\cdot)$:
\begin{align}
M(\beta, \theta)&=W(T(\beta, \theta), J(\beta), \theta, \mathcal{W}), \\
T(\beta, \theta)&={\rm \bar{T}} + B_S(\beta) + B_P(\theta),
\end{align}
where $\mathcal{W}$ refers to the skinning weights. 

In the skinning procedure, not only the hand mesh can be obtained, but also the 3D hand joints can be achieved by rotating joints $J(\beta)$ with the pose parameters $\theta$. 
Thus, an alternative way to estimate 2D hand pose is to project 3D hand joints with proper camera parameters.
Particularly, in this work, we assume an ideal pinhole camera setting, with the projection matrix represented as,
\begin{equation}
\mathbf{Q}=\left(\begin{array}{lll}
f & 0 & p_{0} \\
0 & f & q_{0} \\
0 & 0 & 1
\end{array}\right),
\end{equation}
where $f$ is the focal length and $(p_0 , q_0 )$ is at the image center position, making $f$ the only unknown variable.

Note that, instead of replicate $K$ regressor heads with stand-alone training parameters, we make the $K$ regressors share a global group of training parameters, similar to the strategy adopted in \cite{wei2016convolutional}.

\begin{figure*}
\center
\includegraphics[width=0.9\textwidth]{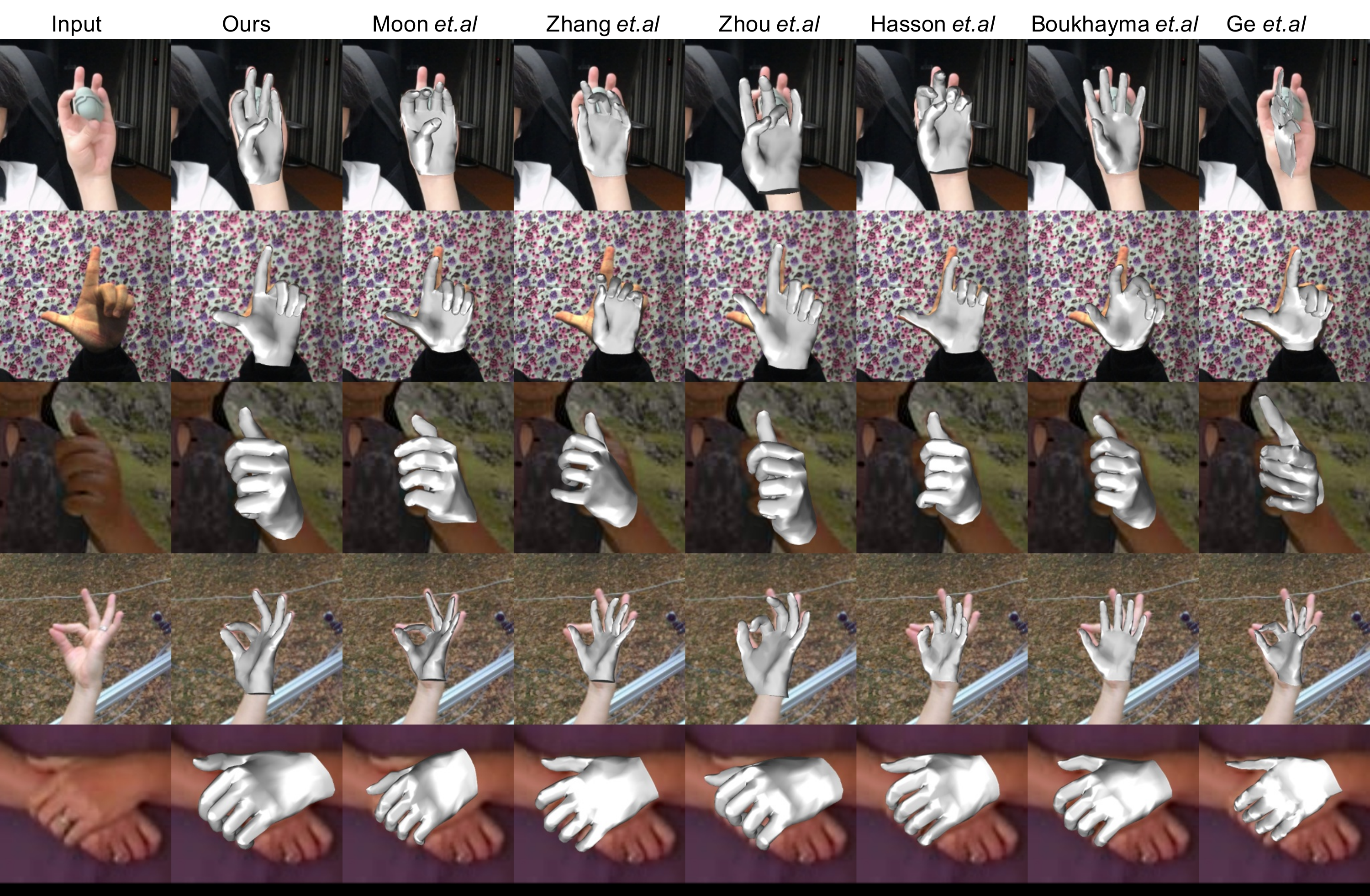}
\caption{{\bf Qualitative Evaluation.} 
The first column exhibits the RGB inputs that cover widely used benchmarks \cite{zhang2017hand,zimmermann2017learning,zimmermann2019freihand,simon2017hand} (2\emph{nd} - 5\emph{th} rows)
 and wild situations (1\emph{st} row). The following columns demonstrate the reconstruction results of ours, 
\cite{moon2020i2l}, \cite{zhou2020monocular}, \cite{zhang2019end}, \cite{hasson2019learning}, \cite{boukhayma20193d}, and \cite{ge20193d} respectively.
To make a fair comparison, all methods are evaluated on the public available pre-trained checkpoints.
}
\label{Fig_Mesh_Recovery_Results}
\end{figure*}

\subsection{Training Objective}
The differentiable property of the backbone, regressor heads and MANO makes our HIU-DMTL framework end-to-end trainable.
The overall loss function can be divided into three categories, targeting at backbone training, the regressor heads optimizing, and self-supervised learning (\emph{loss weights are ignored in the following discussions}). 

\textbf{Training the Backbone.} 
The target of the backbone is to derive certain kinds of instrumental representations from the hand image, including the 2D heat-maps, hand mask, and the POF encoding.
To train the backbone, the outputs of three branches in every MTL block are directly supervisely trained. 
Specifically, the training objective is defined as:
$\mathcal{L}_{backbone} = \mathcal{L}_{hm} + \mathcal{L}_{pof} + \mathcal{L}_{seg}$, 
where $\mathcal{L}_{hm}$ makes the estimated heat-maps close to the ground truth, the $\mathcal{L}_{pof}$ as well. And the $\mathcal{L}_{seg}$ makes use of the classical cross-entropy loss often used in semantic image segmentation task.


\textbf{Training the Regressor Heads.}
The regression module aims to regress the camera parameters $\{R, T\}$ and mesh parameters $\{\beta, \theta\}$ of MANO.
However, it is impracticable to obtain the ground-truth labels.
Fortunately, the regressors can be trained with widely-available samples with 3D/2D annotations through weakly-supervised learning.
Concretely, the loss function comprises three terms, $\mathcal{L}_{regressor}=\mathcal{L}_{3d} + \mathcal{L}_{2d} + \mathcal{L}_{mask}$, 
where $\mathcal{L}_{3d}$ measures the orientation similarity between the estimated bones and the ground-truth bones in aforementioned skeleton hierarchy $\mathbb{S}$,
while $\mathcal{L}_{2d}$ and $\mathcal{L}_{mask}$ refer to the re-projection loss of 2D joints and hand mask, as in \cite{boukhayma20193d,zhang2019end,ge20193d}.

\textbf{Achieving Self-Supervised Learning.}
For a well reconstructed hand mesh, the projected mask shall be consistent with the silhouette, and such constraint has been exploited when training the regressor heads.
However, more implicit constraints shall be maintained among the reasonable predictions.
For instance, the rendered hand mask shall match the mask estimated by the backbone; the coordinates of the re-projected hand pose shall be close to the one inferred from the heat-maps of the backbone. Such consistencies enable us to exploit unlabeled hand images to achieve self-supervised learning.
In practice, the re-projected differentiable hand mask is obtained via differentiable renderers contained in Pytorch3D \cite{ravi2020accelerating},
and Equation \ref{eq:dif_pose} is adopted to derive the differentiable 2D pose from the heat-maps.
\begin{figure}
\centering
\includegraphics[width=0.4\textwidth]{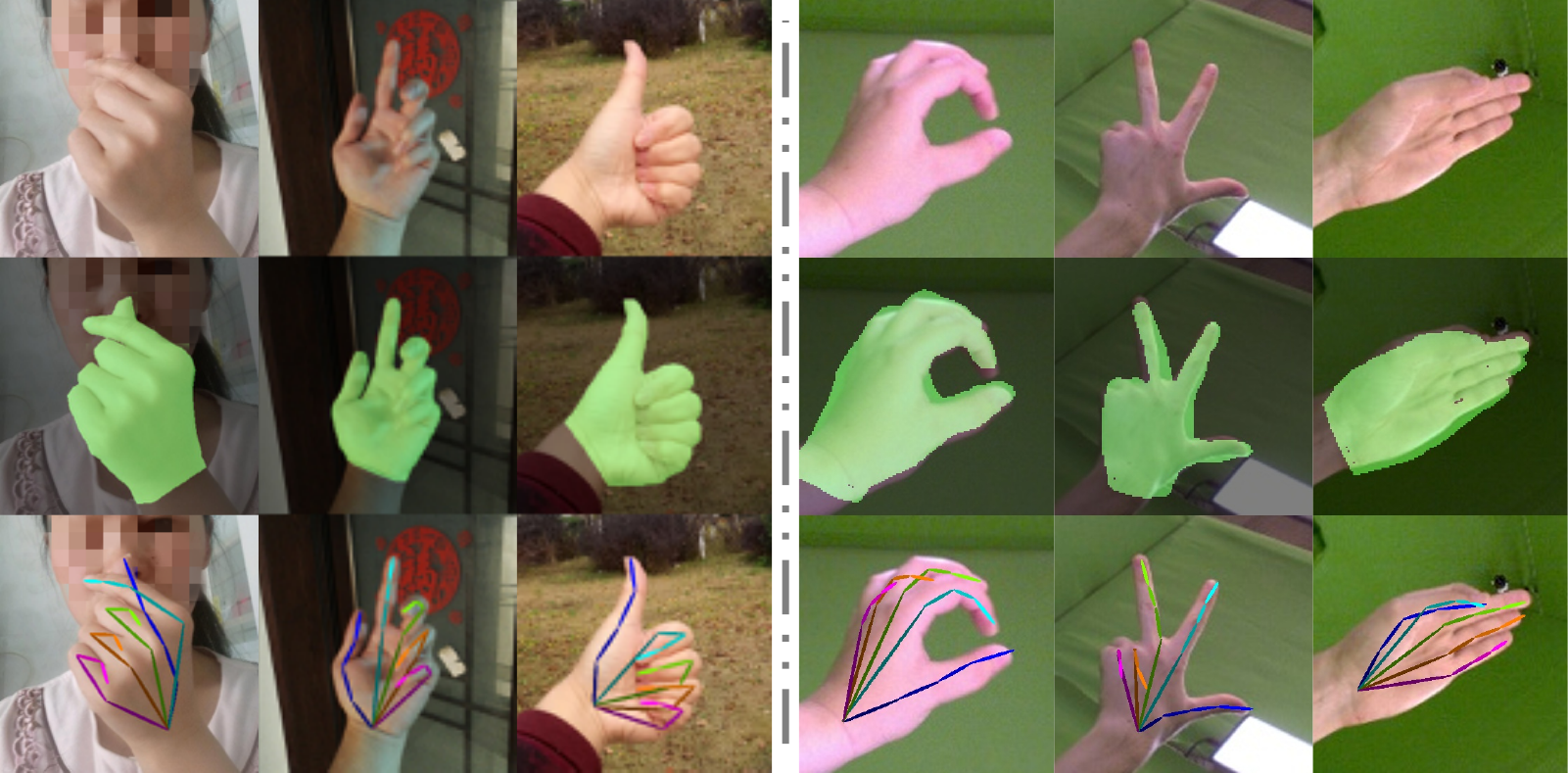}
\caption{{\bf Comparisons between HIU-Data and FreiHAND dataset.}
\label{fig:HIU_Data}
The first three columns demonstrate samples from the HIU-Data, and the last three columns present examples in FreiHAND \cite{zimmermann2019freihand}.
For each sample, the center cropped hand image, the segmentation mask, and the 2D hand pose are visualized.
\vspace{-11pt}
}
\end{figure}

\begin{figure}
\center
\includegraphics[width=0.4\textwidth]{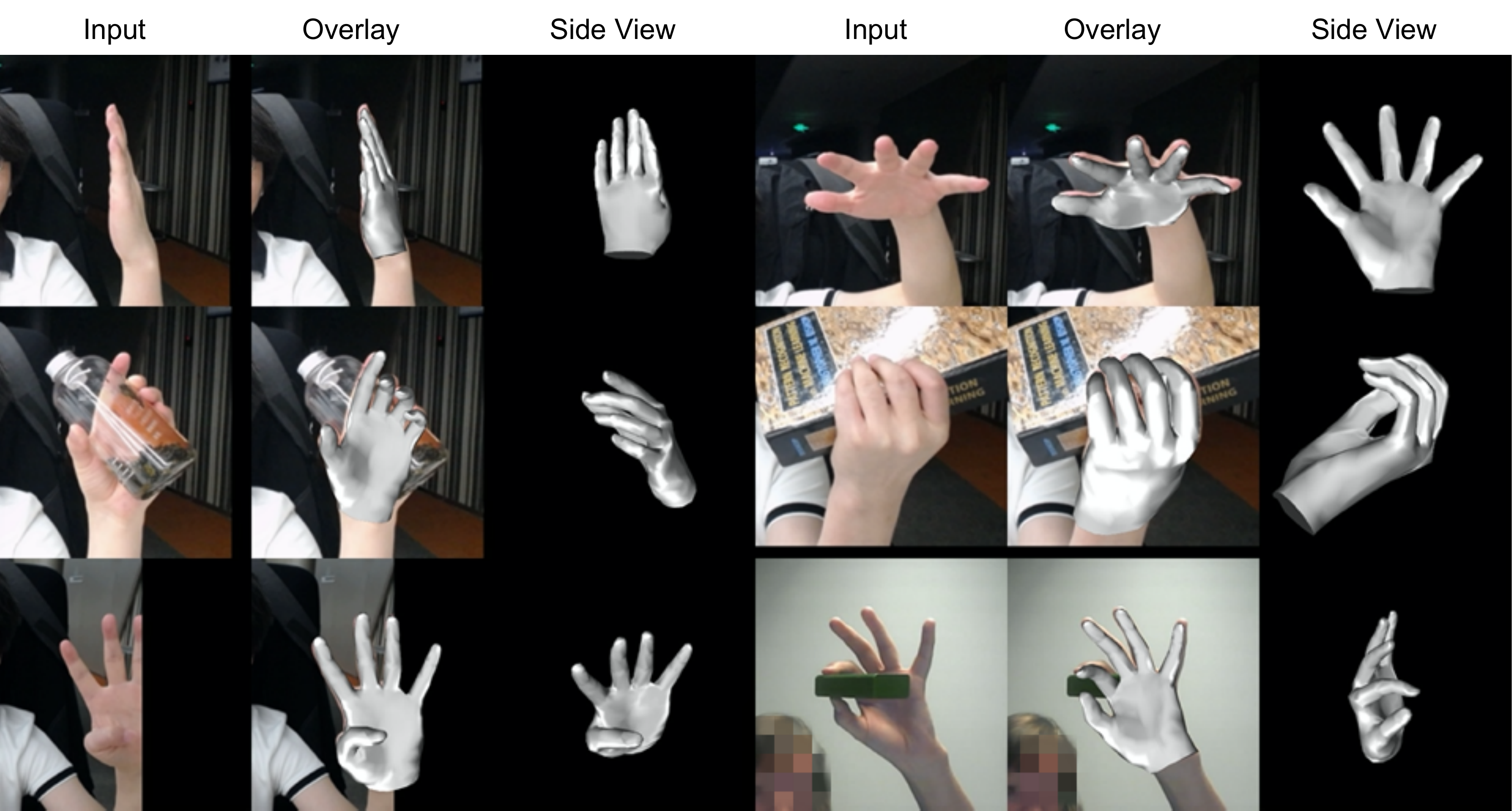}
\caption{{\bf Mesh Recovery Results.} 
We demonstrate the performance of our method under several typical challenging scenarios.
For each sample, the overlay results between reconstructed hand and input image, and the side-view rendering results are presented.
}
\label{Fig_Mesh_Recovery_abs}
\end{figure}

\begin{figure*}
	\centering
	\subfigure[3D PCK on STB \cite{zhang2017hand} ]{\label{3dpose_stb} \includegraphics[width=0.5\columnwidth]{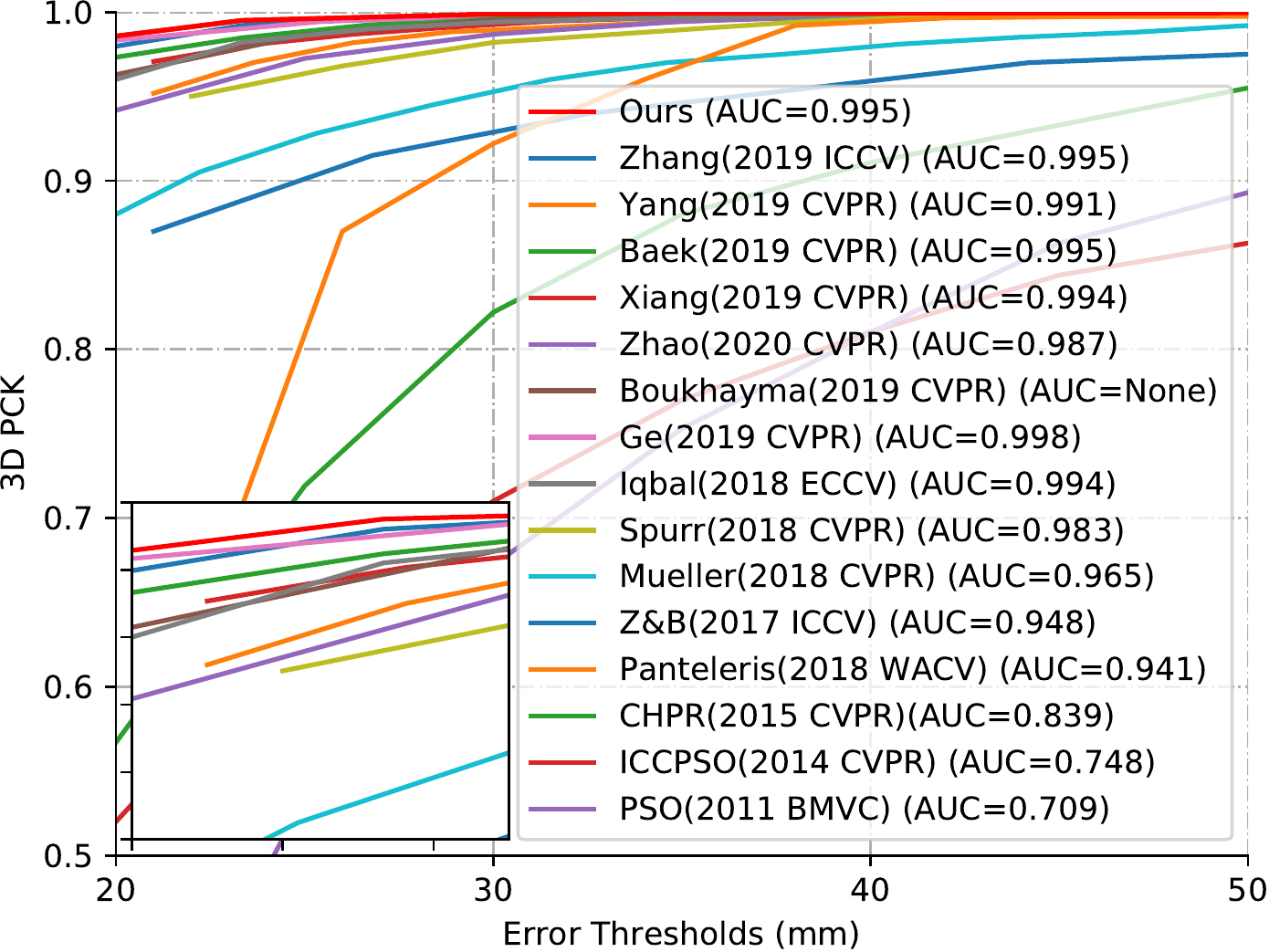}}
	\subfigure[3D PCK on RHD \cite{zimmermann2017learning} ]{\label{3dpose_rhd} \includegraphics[width=0.5\columnwidth]{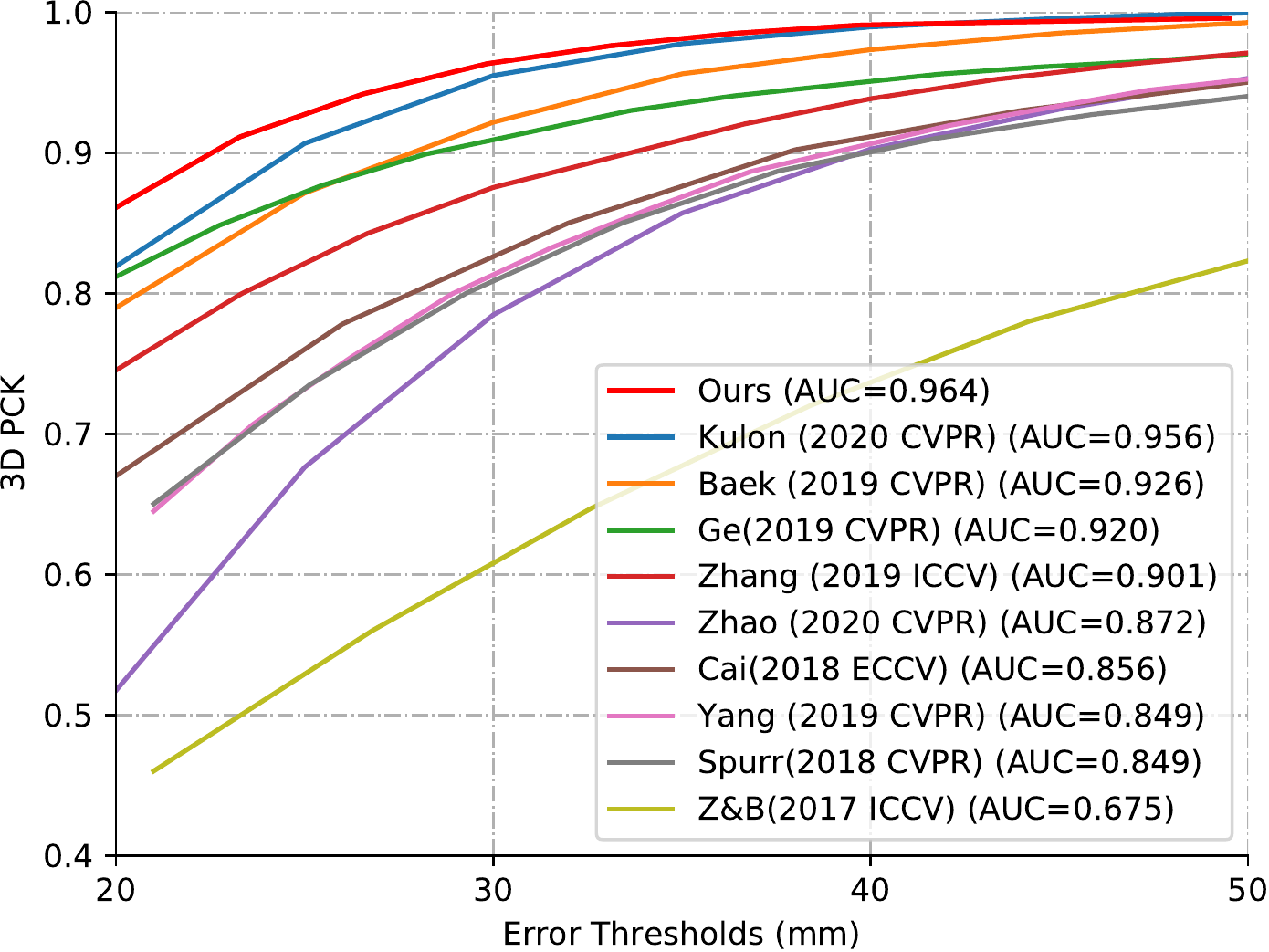}}	
		\subfigure[3D PCK on Dexter \cite{sridhar2016real} ]{\label{3dpose_dexter} \includegraphics[width=0.5\columnwidth]{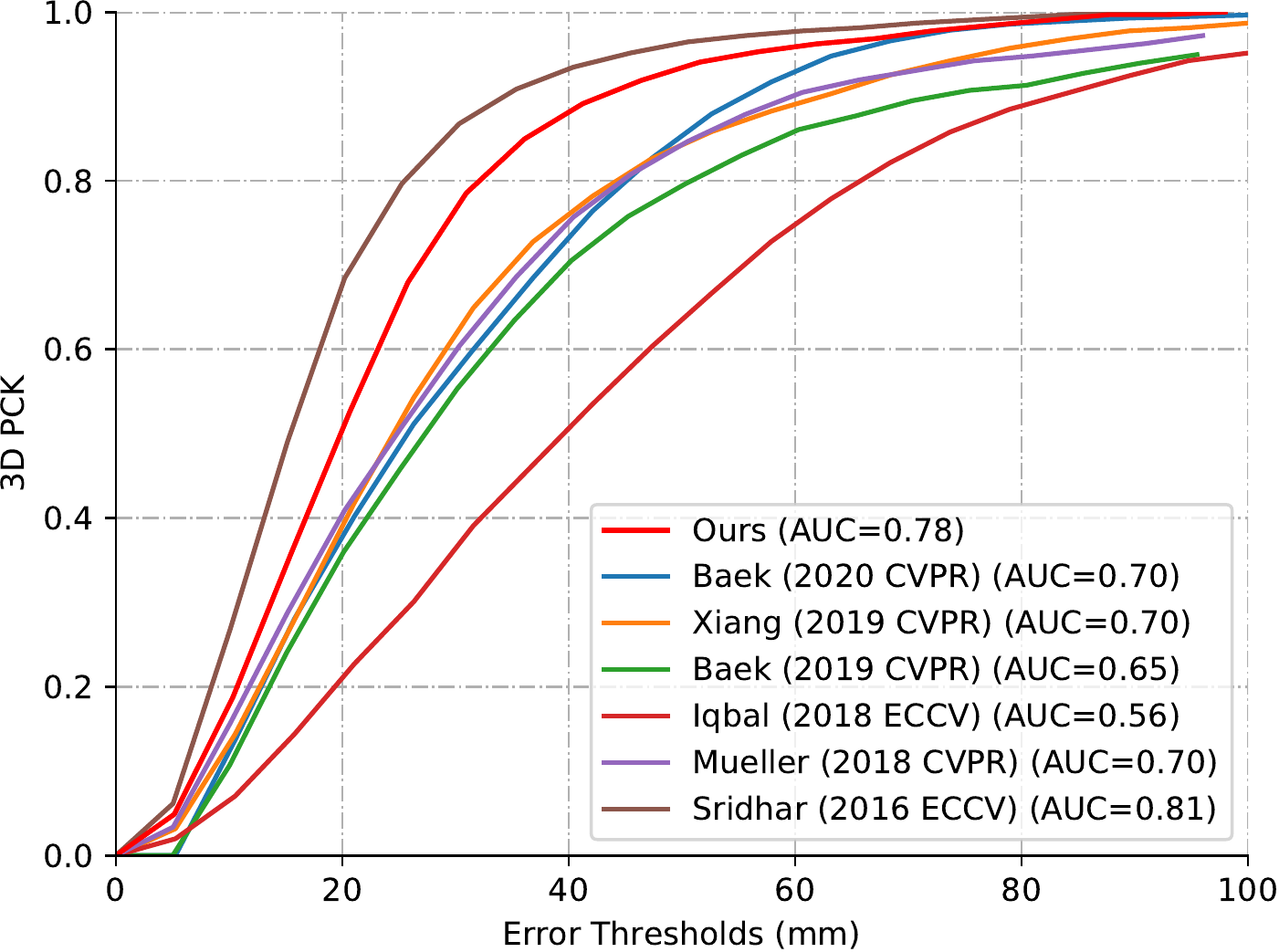}}	
	\subfigure[3D PCK on FreiHAND \cite{zimmermann2019freihand} ]{\label{3dpose_freihand} \includegraphics[width=0.5\columnwidth]{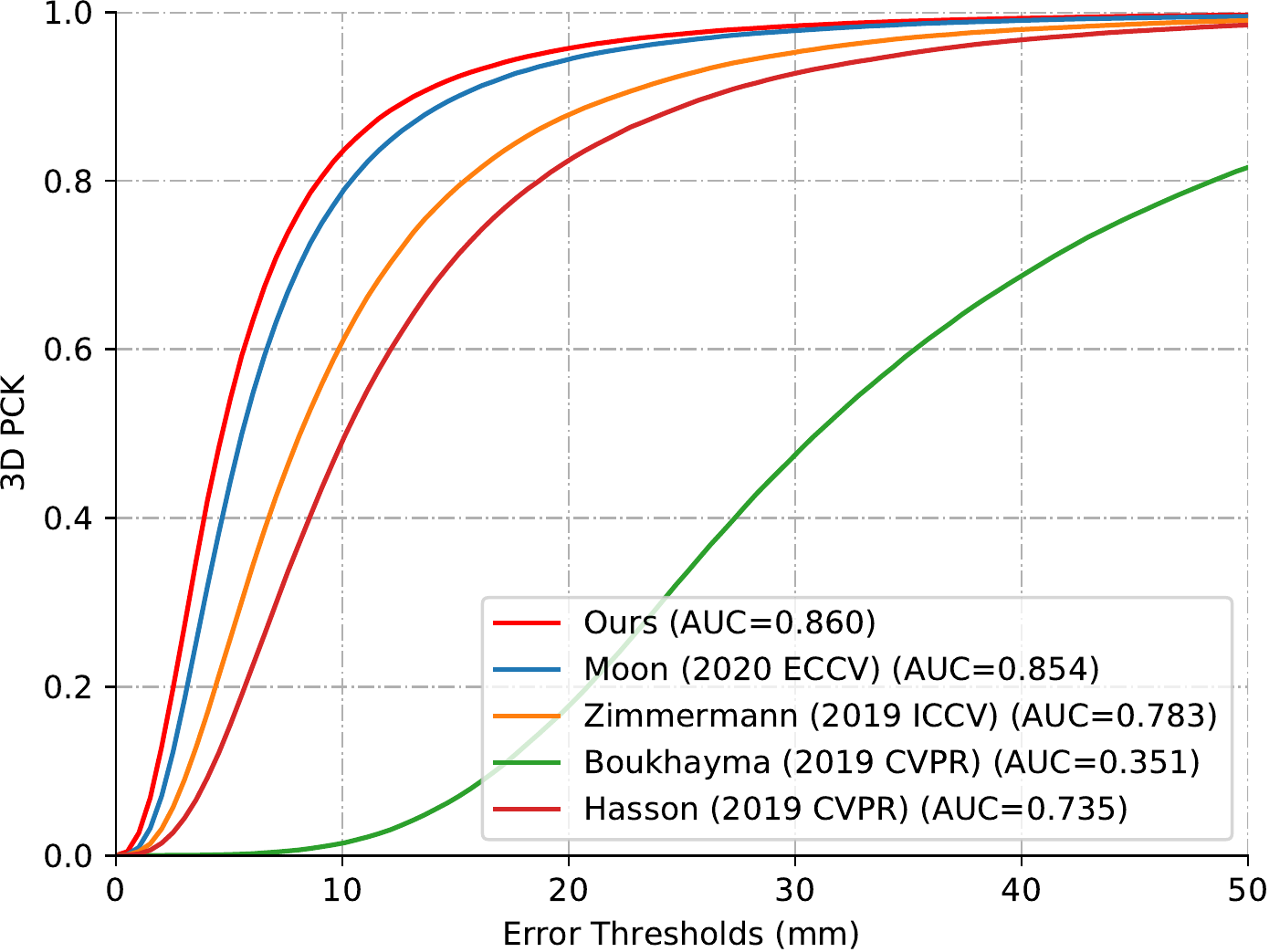}}	
		\caption{\textbf{Quantitative Evaluation.} 
		The plots present the 3D PCK on the STB, RHD, Dexter, and FreiHAND datasets, respectively.
}
\label{fig:quan_3d_pose}
\end{figure*}

\begin{figure*}
	\centering
	\subfigure[2D PCK on HIU-Data]{\label{2dpose_hiu} \includegraphics[width=0.5\columnwidth]{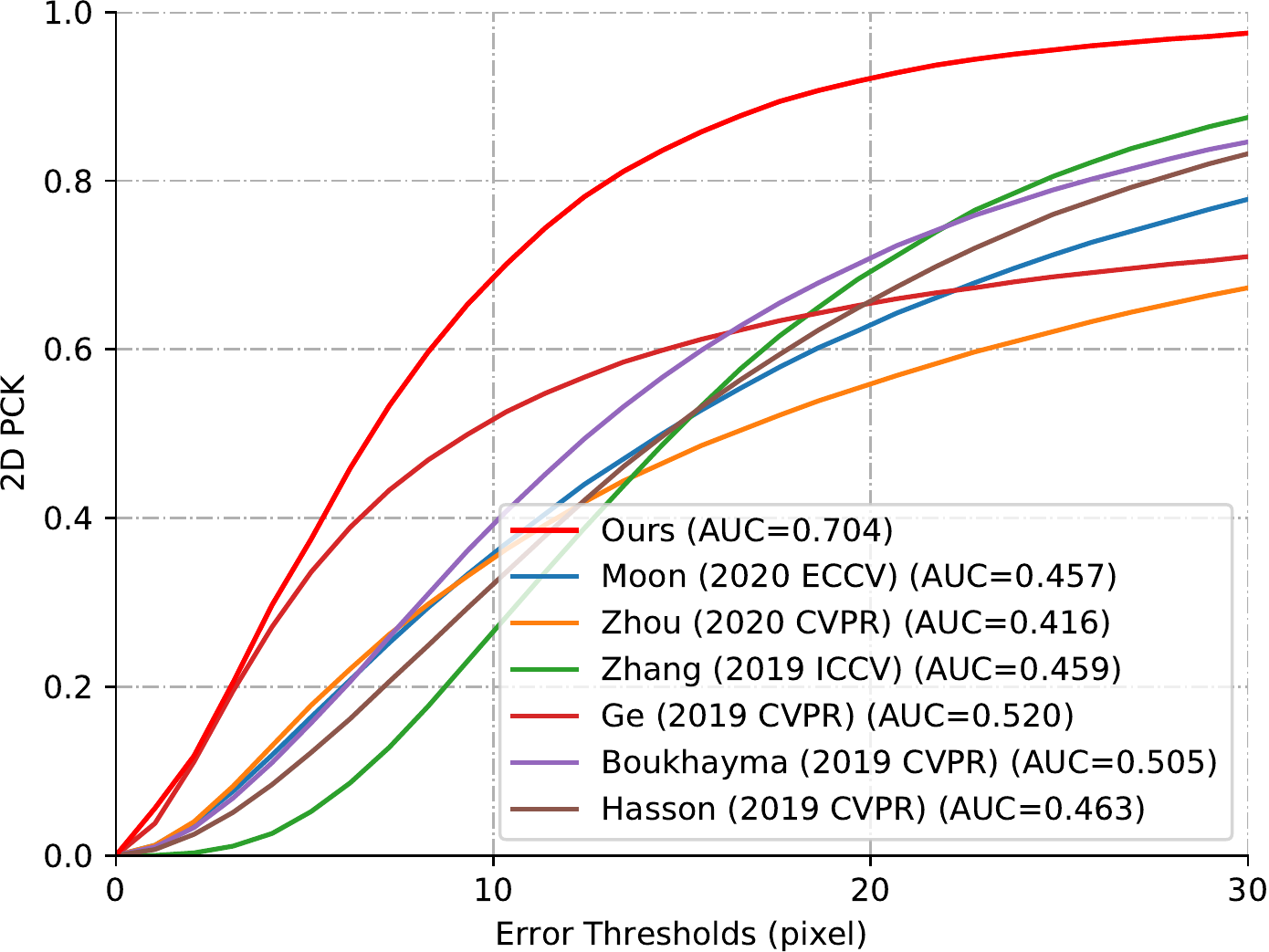}}
	\subfigure[IOU PCK on HIU-Data]{\label{seg_hiu} \includegraphics[width=0.5\columnwidth]{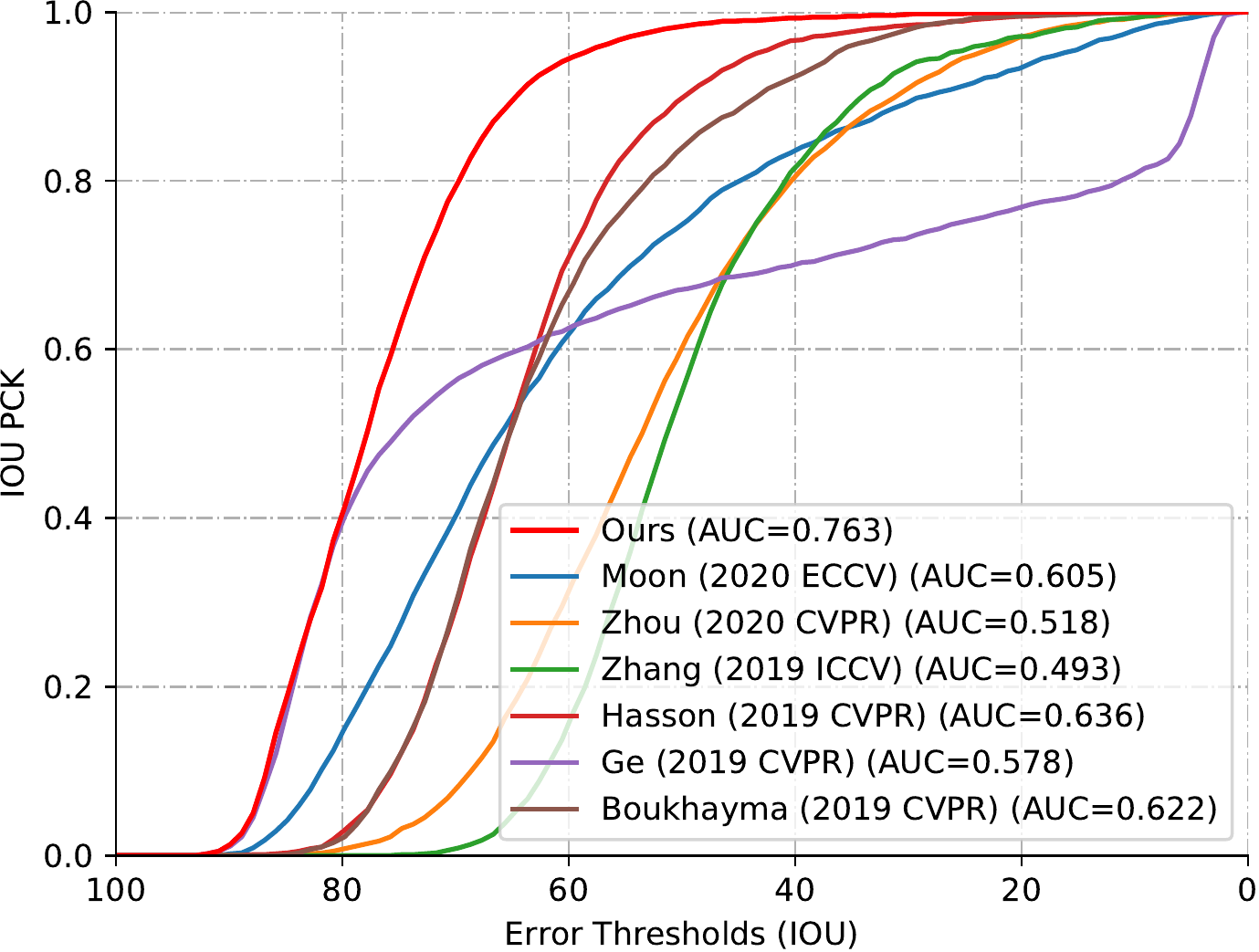}}	
	\subfigure[2D PCK on CMU \cite{simon2017hand}  ]{\label{2dpose_cmu} \includegraphics[width=0.5\columnwidth]{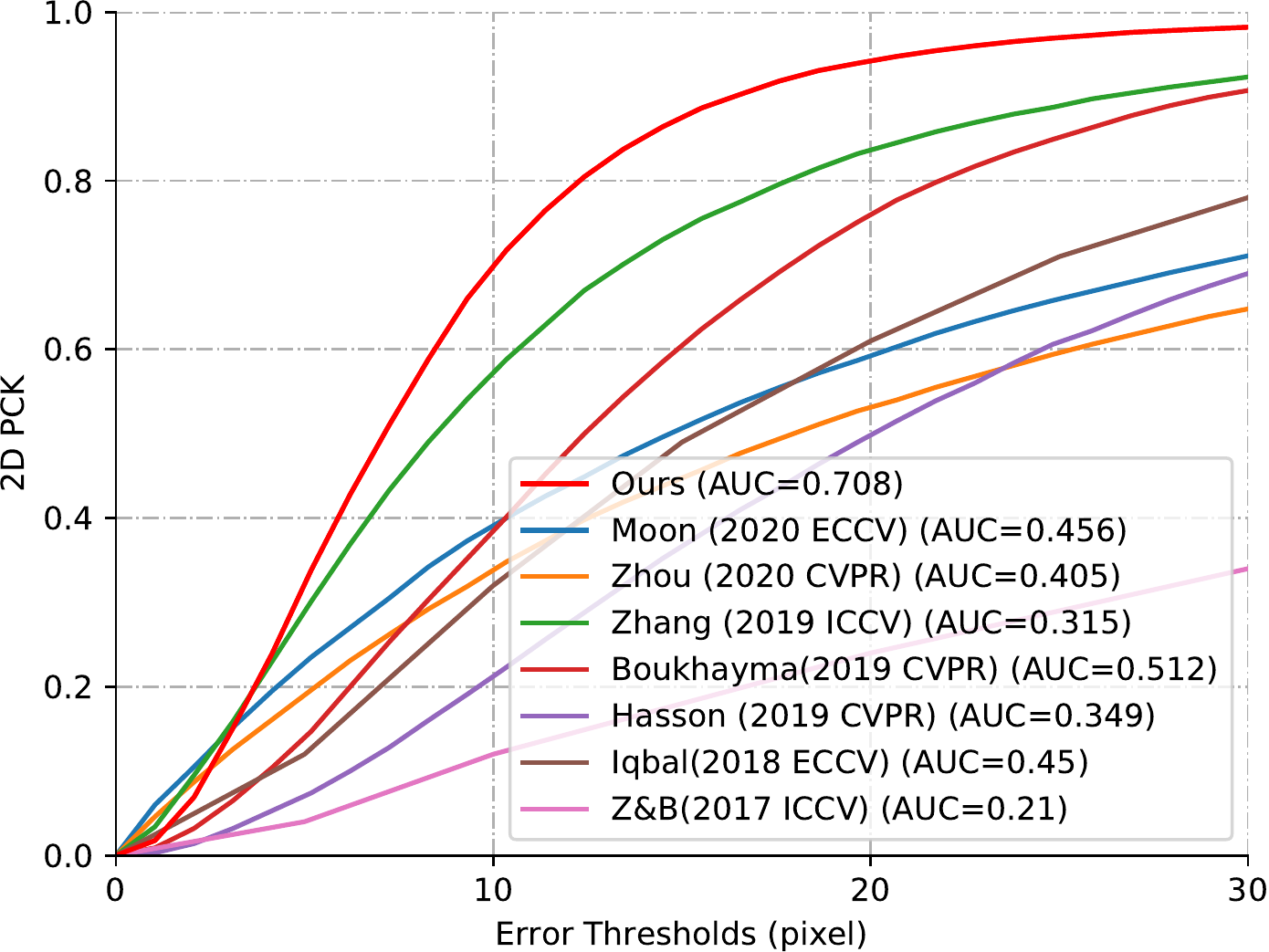}}	
	\subfigure[Mesh PCK on FreiHAND \cite{zimmermann2019freihand} ]{\label{3dmesh_freihand}\includegraphics[width=0.5\columnwidth]{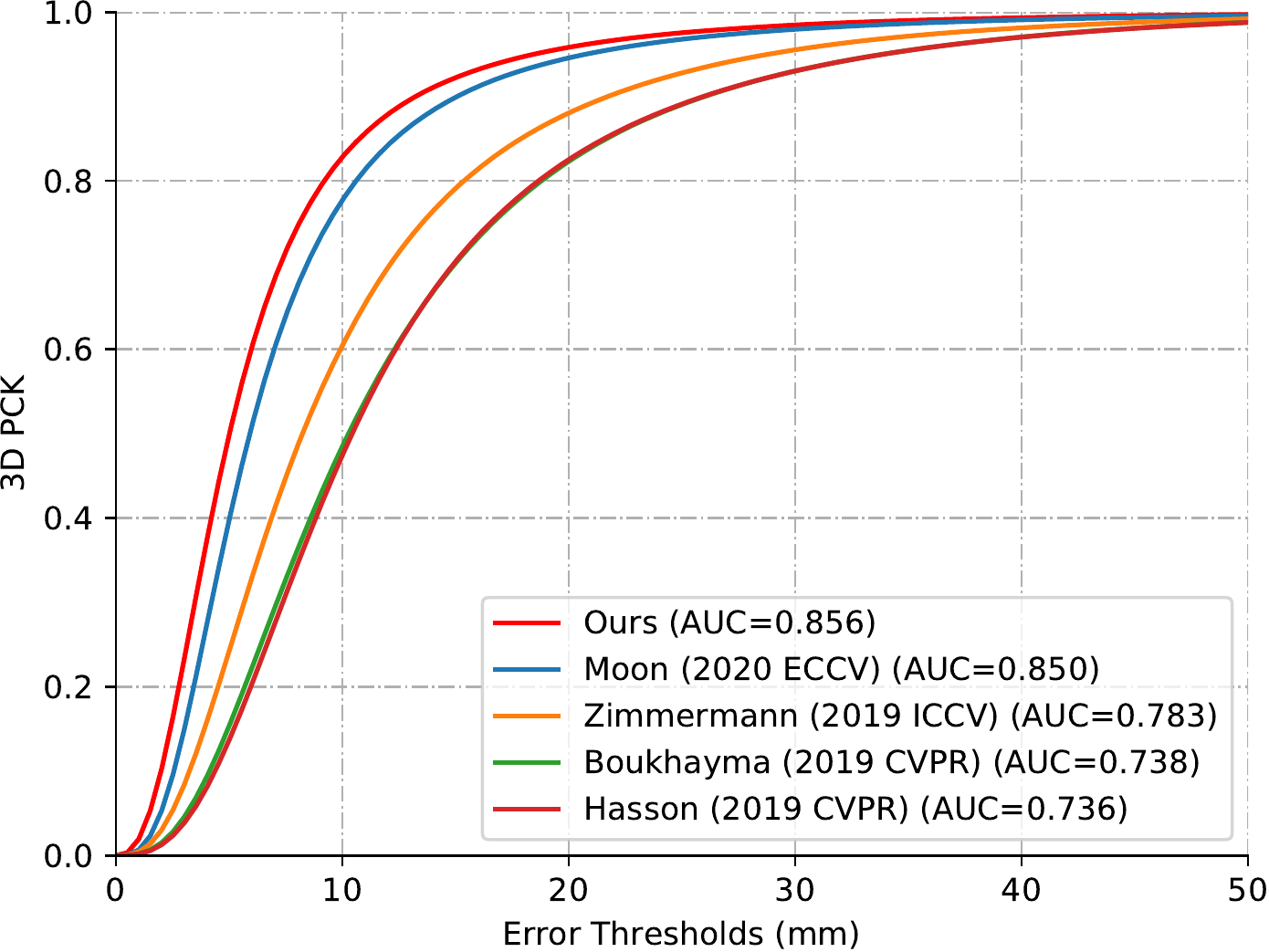}}	
		\caption{\textbf{Quantitative Evaluation.} 
		The figures demonstrate the 2D PCK of the re-projected hand pose on HIU-data, IOU PCK of the re-projected hand mask on HIU-data, 2D PCK of the re-projected hand pose on CMU, 3D per-vertex PCK on FreiHAND, respectively.
}
\label{fig:quan_3d_mesh}
\end{figure*}

\begin{figure}
\centering
\includegraphics[width=0.4\textwidth]{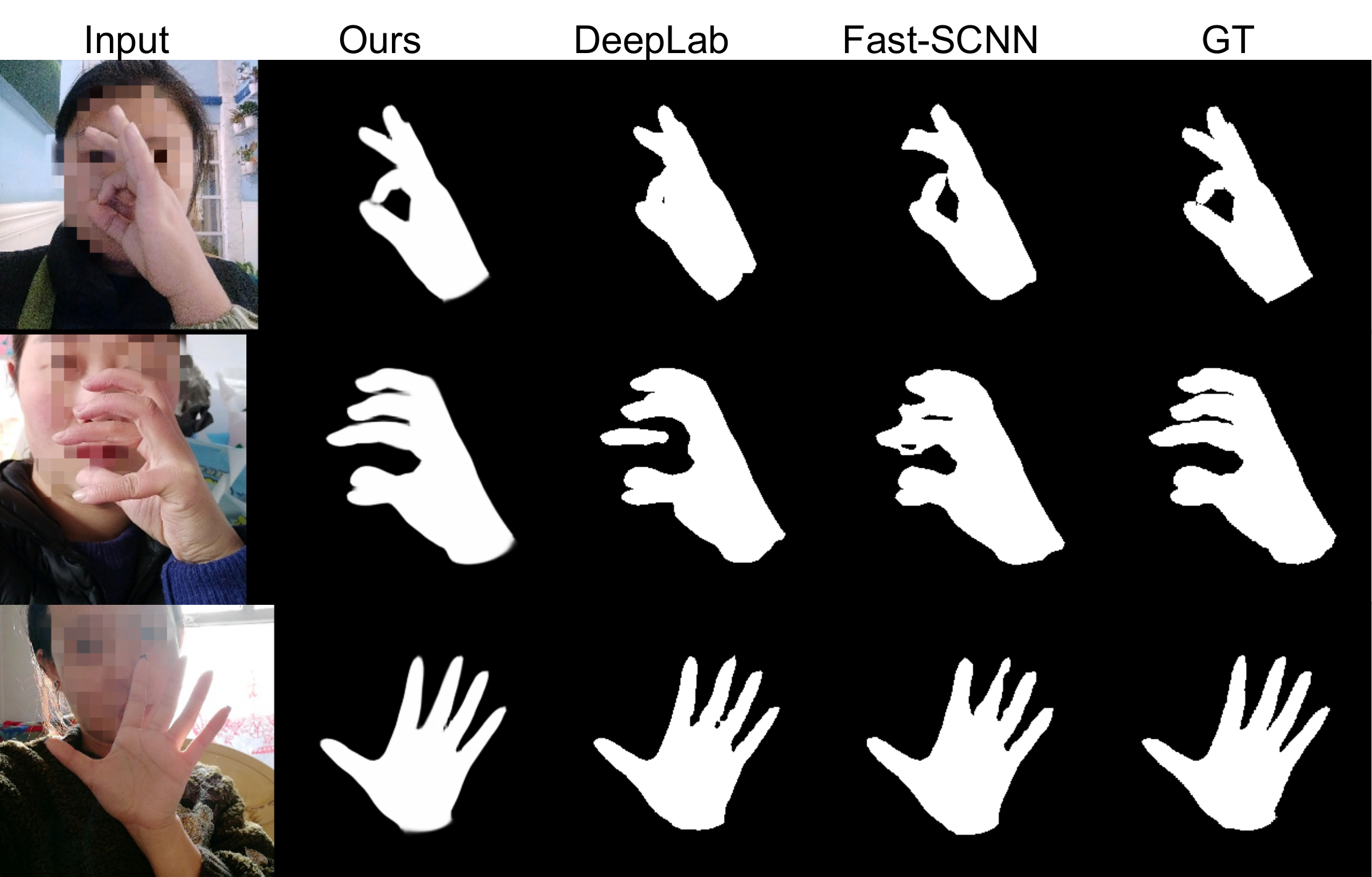}
\caption{{\bf Hand Segmentation.}
The figure presents several results of mask segmentation task from our HIU-DMTL, \cite{chen2017deeplab}, and \cite{poudel2019fast}.
\label{fig:compare_seg}
}
\end{figure}

\section{Experiments}
To demonstrate the effectiveness of our HIU-DMTL framework, we first present qualitative comparison over our recovered hand mesh with the results from recent works.
We then quantitatively evaluate the superiority of HIU-DMTL on 2D/3D hand pose estimation, hand mask segmentation, and hand mesh reconstruction tasks over several publicly available datasets.
Finally, we perform several ablation studies to evaluate the importance of different design strategies and analyze the impact of the hyper-parameters. 
Due to the limited space, the implementation details of experiments are provided in the supplemental materials, please refer it for more details.

\subsection{Experiment Settings}
\textbf{Datasets.}
We mainly leverage two kinds of datasets, \emph{i.e.,} publicly available benchmarks and our newly annotated dataset.
For public datasets, we evaluate our method on CMU Panoptic Dataset (CMU) \cite{simon2017hand}; the Rendered Hand dataset (RHD) \cite{zimmermann2017learning}; the Stereo Hand Pose Tracking Benchmark (STB) \cite{zhang2017hand}; the FreiHAND \cite{zimmermann2019freihand}; and the Dexter Object (Dexter) \cite{sridhar2016real} dataset.
Since no accessible datasets contain high-quality hand masks, making it difficult to perform training directly on aforementioned HIU tasks. 

\textbf{HIU-Data.} Regarding above concern, we manually collect 33,000 hand images in challenging scenarios to be our new dataset, namely HIU-data, 
For each sample, both the 2D hand pose and hand mask are manually annotated rather than generating approximate labels automatically as did in \cite{zimmermann2019freihand, xiang2019monocular, yu2020humbi}. 
As Figure \ref{fig:HIU_Data} demonstrates several samples with the corresponding labels, one may observe that our dataset achieves better annotation accuracy than the newest released automatically annotated FreiHAND \cite{zimmermann2019freihand} dataset.


\subsection{Qualitative Evaluation}
To visually evaluate the quality of the recovered hand mesh representation and the robustness of HIU-DMTL in various cases, we illustrate several representative samples in Figure \ref{Fig_Mesh_Recovery_Results} and Figure \ref{Fig_Mesh_Recovery_abs}.
As shown in Figure \ref{Fig_Mesh_Recovery_Results}, the recovered hand mesh of HIU-DMTL have better overlays with the input image than the those reconstructed by other contemporary approaches \cite{moon2020i2l, zhou2020monocular, zhang2019end,hasson2019learning, boukhayma20193d, ge20193d}.
Furthermore, HIU-DMTL demonstrates superior robustness and generalization capability on in-the-wild unseen images, which is infeasible for previous existing methods. 
For example, Figure \ref{Fig_Mesh_Recovery_abs} shows that our method is robust enough to conduct hand meshes recovery in several challenging situations, such as exaggerated hand articulations, extremely unconstrained camera views, the existence of image truncations, heavy self-occlusions, and hand object interaction.

\begin{table}
\centering
{
\resizebox{0.4\textwidth}{!}
{%
\begin{tabular}{l | c c c}
\toprule
Method& DeepLab v3 \cite{chen2017deeplab} & Fast-SCNN \cite{poudel2019fast} &  HIU-DMTL \\
\hline
mIoU $\uparrow$ & 96.6\%  & 96.2\%   &97.5\% \\
\bottomrule
\end{tabular}
}
}
\vspace{3pt}
\caption{\textbf{Hand Segmentation.} The table presents the mIoU of \cite{chen2017deeplab}, \cite{poudel2019fast}, and our HIU-DMTL on HIU-Data.
}
\label{tab:quan_seg}
\end{table}

\subsection{Quantitative Evaluation}
We quantitatively evaluate the superiority of HIU-DMTL on 2D/3D hand pose estimation, hand mask segmentation, and hand mesh reconstruction tasks respectively.

\textbf{Comparisons of 3D Hand Pose.} To be consistent with \cite{boukhayma20193d,zhang2019end,hasson2019learning,ge20193d,zhou2020monocular}, 3D PCK is adopted to evaluate the performance of 3D hand pose estimation.
All the comparison methods are evaluated on STB, RHD, Dexter, and FreiHAND datasets, and Figure \ref{fig:quan_3d_pose} reports the overall experimental results.
On the \textbf{STB} dataset, the 3D PCK curves remain intertwined because the STB dataset is relatively small and lacks diversity. 
Our method achieves competitively results among newly launched approaches
\cite{boukhayma20193d,zhang2019end,ge20193d,yang2019disentangling,baek2019pushing,xiang2019monocular,zhao2020knowledge}, which is reasonable when considering the saturated performance on this dataset \cite{zhang2019end,zhou2020monocular,xiang2019monocular,yang2019aligning,iqbal2018hand}.
On the \textbf{RHD} dataset, which is relatively complex and more diverse, our method outperforms all existing approaches \cite{kulon2020weakly,zhang2019end,baek2019pushing,ge20193d,yang2019disentangling,zhao2020knowledge} and achieves state-of-the-art performance.
On the \textbf{Dexter Object} dataset, our method dramatically outperforms existing works \cite{baek2020weakly,xiang2019monocular,baek2019pushing,iqbal2018hand,mueller2018ganerated} and remains comparable with \cite{sridhar2016real}, which exploits additional depth information to calibrate the joint locations. 
On the \textbf{FreiHAND} benchmark, our approach outperforms contemporary methods \cite{hasson2019learning,boukhayma20193d,zimmermann2019freihand} by a large margin, and obtains slightly better performance comparing with \cite{moon2020i2l}, while \cite{moon2020i2l} exploits the additional ground truth hand mesh to fulfill mesh recovery.

\textbf{Comparisons of Hand Mesh.} We perform apple to apple comparison with these approaches \cite{moon2020i2l,zhou2020monocular,zhang2019end,ge20193d,boukhayma20193d,hasson2019learning}, containing accessible source code/checkpoint, which are focused on hand mesh recovery task under various evaluation metrics.
Specifically, we take the IoU PCK of the re-projected mask, the 2D PCK of re-projected hand pose, and the 3D PCK of per-vertex reconstruction error as the evaluation metrics, and evaluate the aforementioned methods on the HIU-Data, CMU, and FreiHAND, as shown in Figure \ref{fig:quan_3d_mesh}.
Regarding the \textbf{2D PCK} of the re-projected 2D hand pose, our method outperforms \cite{moon2020i2l,zhou2020monocular,zhang2019end,ge20193d,boukhayma20193d,hasson2019learning} by a wide margin.
Regarding the \textbf{IOU PCK} of the re-projected hand mask, our approach retains a significantly higher PCK score, which is also already demonstrated in Figure \ref{Fig_Mesh_Recovery_Results} that our recovered hand mesh overlays with the input images better. 
Regarding the \textbf{3D PCK} of per-vertex reconstruction error, our framework obtains the minimum misalignment, comparing to  \cite{moon2020i2l,zimmermann2019freihand,boukhayma20193d,hasson2019learning}. 
Note that the method \cite{moon2020i2l} additionally exploits the ground truth hand mesh in FreiHAND dataset for training.

\textbf{Comparisons of Hand Mask.} 
For the hand mask segmentation, we compare our approach with general SOTA segmentation methods \cite{chen2017deeplab,poudel2019fast}.
Table \ref{tab:quan_seg} reveals that our method is more suitable than general sophisticated approaches for the task of hand mask segmentation.
As the certain representative samples shown in Figure \ref{fig:compare_seg}, our approach can estimate a precise mask even in the texture area with high-frequency, which is incapable in \cite{chen2017deeplab,poudel2019fast}.
Besides, \cite{chen2017deeplab,poudel2019fast} may generate unreasonable hand masks that violates the biomedical appearance of hand. 
However, our approach can avoid such a dilemma by fully exploiting SSL and using priors that are comprised in hand mesh.

\subsection{Ablation Study}
The superior performance of HIU-DMTL mainly attributes to the cascaded design (CD) paradigm, the multi-task learning (MTL) setup, the task-attention module (TAM), and the self-supervised learning (SSL) strategy. In this section, we perform ablation studies to better understand the importance of these different design choices.
Also, the ablation studies deeply analyze the impact of some parameters related with network structure, such as the number of the network stack in our framework.

\textbf{Cascaded Design Paradigm.} The CD paradigm has been widely adopted in 2D pose estimation task \cite{newell2016stacked,wei2016convolutional, sun2015cascaded, tang2015opening}, while \textbf{the multi-branched cascaded MTL design for HIU tasks remains less explored}. 
Similar to \cite{newell2016stacked}, we investigate the impact of the number of MTL blocks.
Specifically, we construct three models, which contain 1, 2, and 4 MTL blocks respectively. 
Meanwhile, the three models are designed to have similar FLOPs for a fair comparison.
As shown in Table \ref{tab:abs_Cascade}, changing from 1 stack to 4 stacks, the performance of 2D hand pose and hand mask tasks can improve substantially.
We also investigate how performance improves, according to the increases of MTL block. Table \ref{tab:abs_per_stack} presents the performance of each intermediate stack in a 8-stack HIU-DMTL framework, where we take same evaluation metrics as in Table \ref{tab:abs_Cascade}.
We can conclude that from stack 1 to stack 4, the performance increases rapidly, and then this growth trend gradually flattened in later stacks. 

\begin{table}
\centering
{
\resizebox{0.48\textwidth}{!}
{%
\begin{tabular}{c | c c | c c | c c}
\toprule
& 2D Pose & 2D Pose$^\dag$ & Hand Mask & Hand Mask$^\dag$ & 3D Pose & Hand Mesh \\
\hline
4-Stack& \textbf{0.866} & \textbf{0.704} & \textbf{0.974} & \textbf{0.770} & \textbf{0.860} & \textbf{0.856} \\
2-Stack& 0.857 & 0.701 & 0.969 & 0.765 & 0.857 & 0.853 \\
1-Stack& 0.837 & 0.686 & 0.948 & 0.752 & 0.852 & 0.842 \\
\bottomrule
\end{tabular}
}
}
\caption{\textbf{Ablation of CD Paradigm.} The table presents the ablation results across different stacking arrangements under various evaluation metrics, where $^\dag$ indicates inferring the pose/mask by projecting the 3D pose/mesh with proper camera parameters.
The 3D hand pose/mesh are quantified on FreiHAND benchmark, while the 2D pose/mask are evaluated on the HIU-Data, since the quality of masks in FreiHAND benchmark is not good enough.
}
\label{tab:abs_Cascade}
\end{table}

\begin{table}
\centering
{
\resizebox{0.48\textwidth}{!}
{%
\begin{tabular}{c | c c | c c | c c}
\toprule
Stack & 2D Pose & 2D Pose$^\dag$ & Hand Mask & Hand Mask$^\dag$ & 3D Pose & Hand Mesh \\
\hline
Stack-1& 0.818 & 0.661 & 0.915 & 0.712 & 0.824 & 0.819 \\
Stack-2& 0.855 & 0.685 & 0.954 & 0.744 & 0.845 & 0.834 \\
Stack-3& 0.865 & 0.700 & 0.968 & 0.764 & 0.854 & 0.847 \\
Stack-4& {0.867} & {0.704} & {0.975} & {0.769} & {0.857} & {0.853} \\
\hdashline
Stack-5& 0.867 & 0.706 & 0.975 & 0.771 & 0.859 & 0.855 \\
Stack-6& 0.868 & 0.706 & 0.976 & \textbf{0.773} & 0.860 & 0.857 \\
Stack-7& 0.869 & 0.705 & 0.976 & 0.771 & \textbf{0.861} & 0.857 \\
Stack-8& \textbf{0.870} & \textbf{0.707} & \textbf{0.977} & \textbf{0.773} & \textbf{0.861} & \textbf{0.859} \\
\bottomrule
\end{tabular}
}
}
\caption{\textbf{Ablation of Intermediate Stacks.} The table presents the performance of the intermediate stacks under various evaluation metrics, where $^\dag$ shares similar definition as in Table \ref{tab:abs_Cascade}.
}
\label{tab:abs_per_stack}
\end{table}

\textbf{Multi-Task Learning Setup.}
Table \ref{tab:abs_MTL_strategy} reports the influences of jointly learning the hand mask segmentation, 2D hand pose estimation, and hand mesh recovery tasks.
One may observe two conclusions: first, when jointly training of any two tasks together, each task's performance is better than training any stand-alone task; second, when jointly learning the whole three tasks, each task's performance is better than when training in any other configurations. 
This is because the aforementioned relationship constraints among elementary tasks can improve performance with each other. For instance, the priors comprised in the hand mesh can correct the unreasonable estimations of hand mask/pose that violate the hand's biomedical appearance. 

\begin{table}
\centering
{
\resizebox{0.48\textwidth}{!}
{%
\begin{tabular}{ccc|cccc}
\toprule
2D Pose & Hand Mask & Hand Mesh     & 2D Pose & Hand Mask & 3D Pose & Hand Mesh \\
\midrule
\yes & \no & \no & 0.752 & - & - & - \\
\no & \yes & \no & - & 0.863 & - & - \\
\no & \no & \yes & - & - & 0.802 & 0.801 \\
\yes & \yes & \no & 0.791 & 0.875 & - & - \\
\no & \yes & \yes & - & 0.891 & 0.816 & 0.813 \\
\yes & \no & \yes & 0.795 & - & 0.812 & 0.808 \\
\hdashline
\yes & \yes & \yes & \textbf{0.802} & \textbf{0.907} & \textbf{0.821} & \textbf{0.817} \\
\bottomrule
\end{tabular}
}
}
\caption{\textbf{Ablation of MTL Setup.} The table presents the comparisons of different task combinations on FreiHAND dataset.
}
\label{tab:abs_MTL_strategy}
\end{table}

\textbf{Task Attention Module.}
To investigate the effectiveness of the TAM, we conduct ablation experiments on 2D hand pose estimation, hand mask segmentation tasks over the HIU-Data, as well as 3D hand pose regression, hand mesh recovery tasks on the FreiHAND benchmark. 
For the baseline model without the TAM, we employ a concatenation followed by a $3\times3$ convolution to aggregate the feature-maps from individual tasks.
As shown in Table \ref{tab:abs_TAM}, we compare the baseline with and without TAM module in 1-stack and 4-stacks setting. As we can see, the TAM significantly improves the performance of individual tasks in both single-stack mode and multi-stacks mode, which means TAM is not only beneficial to the aggregation of feature representations for regressor branch but also helpful to the transition of feature representation between cascaded stacks. 

\begin{table}
\centering
{
\resizebox{0.48\textwidth}{!}
{%
\begin{tabular}{l | c c | c c | c c}
\toprule
& 2D Pose & 2D Pose$^\dag$ & Hand Mask & Hand Mask$^\dag$ & 3D Pose & Hand Mesh \\
\hline
4-stack           & \textbf{0.866} & \textbf{0.704} & \textbf{0.974} & \textbf{0.770} & \textbf{0.860} & \textbf{0.856} \\
4-stack$^\S$ & 0.793 & 0.662 & 0.957 & 0.739 & 0.813 & 0.808 \\
\hdashline 
1-stack           & 0.837 & 0.686 & 0.948 & 0.752 & 0.852 & 0.842\\
1-stack$^\S$  & 0.821 & 0.648 & 0.935 & 0.739 & 0.841 & 0.826\\
\bottomrule
\end{tabular}
}
}
\caption{\textbf{Ablation of TAM.} The table presents the ablation results of the TAM under various evaluation metrics, where $^\S$ denotes to not employ the TAM, and $^\dag$ shares similar definition as in Table \ref{tab:abs_Cascade}.
}
\label{tab:abs_TAM}
\end{table}

\begin{figure}
	\centering
	\subfigure[2D AUC on HIU-Data ]{\label{abs_2dpose_hiu} \includegraphics[width=0.48\columnwidth]{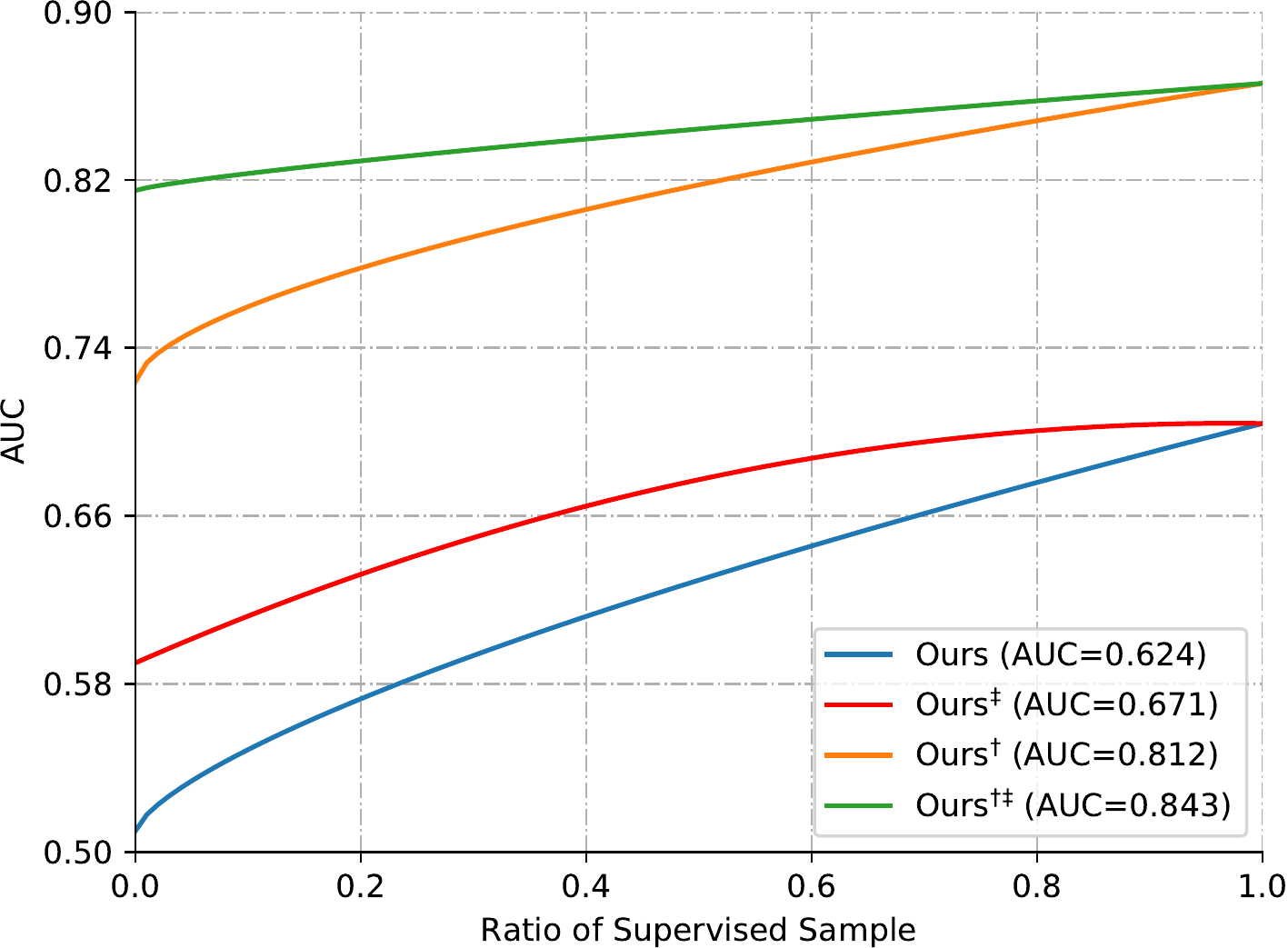}}
	\subfigure[IOU AUC on HIU-Data  ]{\label{abs_seg_hiu} \includegraphics[width=0.48\columnwidth]{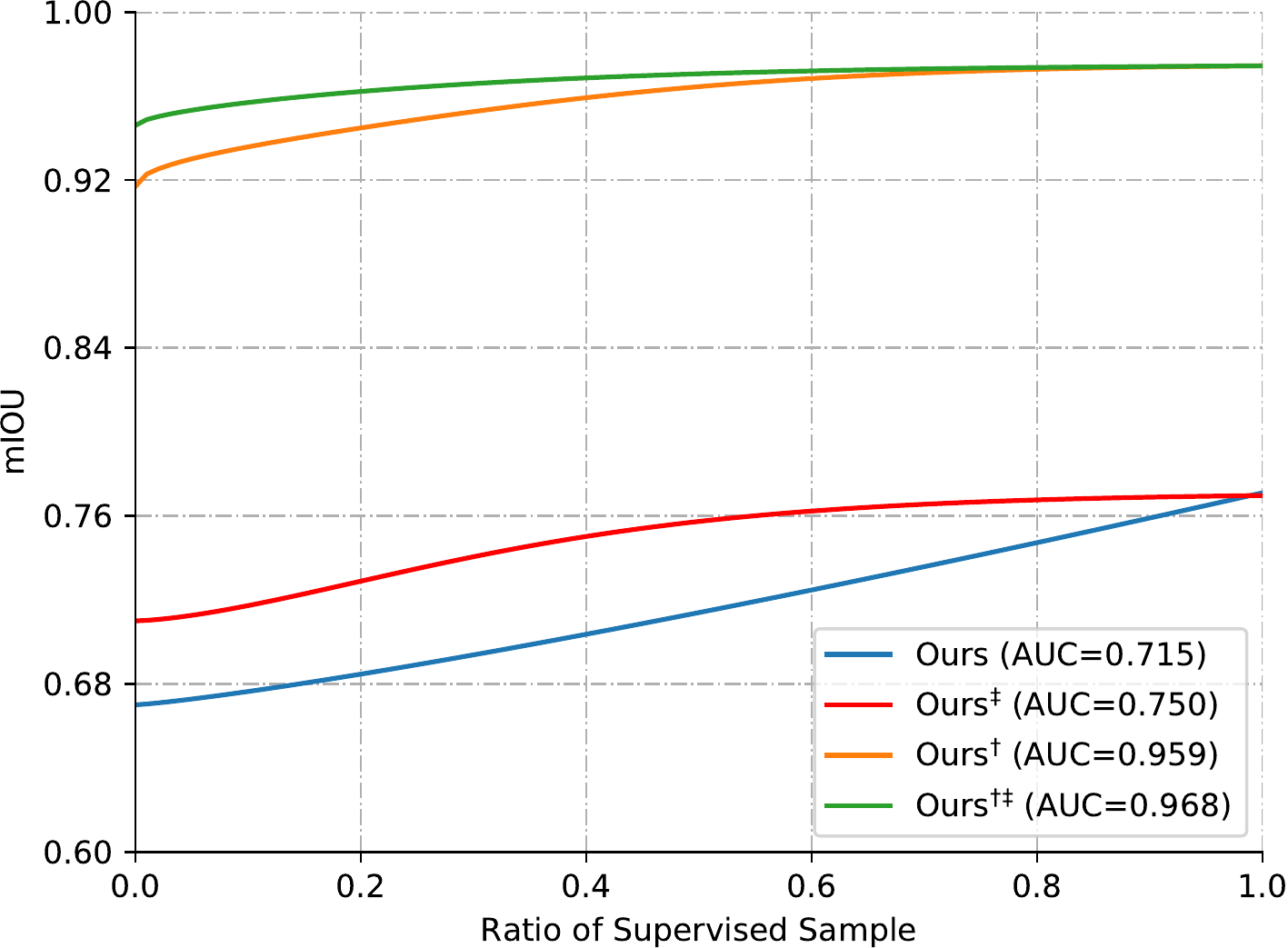}}	

\caption{\textbf{Ablation of SSL Strategy.} 
The plots present the comparison of with and without applying self-supervised learning strategy over the various portions of samples that are used for supervised learning, $^\ddag$ denotes to apply self-supervised training over the remaining samples that are not chosen to do supervised learning,
$^\dag$ shares the similar definition as in Table \ref{tab:abs_Cascade}.
}
\label{fig:abs_self_supervised_pose}
\end{figure}

\textbf{Self-Supervised Learning Strategy.} To verify the effectiveness of the SSL paradigm, we take the AUC of 2D hand pose and hand mask as the evaluation metric and compare the performance of framework with and without SSL. 
As plotted in Figure \ref{fig:abs_self_supervised_pose} (left), the 2D hand pose retrieved from the backbone and 2D hand pose that derived from projecting the 3D hand joints are two factors evaluated in this comparison. 
One may observe that: when no samples are used for supervised training (x=0), the self-learning strategy can 
improve baseline performance dramatically. 
As all samples are under supervised training mode (x=1), the vacancy of self-supervised learning results in no influence over final performance at all. 
Similarly, the right side of Figure \ref{fig:abs_self_supervised_pose} plots the comparisons of hand mask, which also demonstrates that SSL strategy can significantly improve the performance of hand mask.
The above observations also explain the generalization capability of HIU-DMTL on unseen in the wild images (Figure \ref{Fig_Mesh_Recovery_abs}) from another aspect.


\vspace{4pt}
\section{Discussion}
In this work, we proposed a novel cascaded multi-task learning (MTL) framework, namely \textbf{HIU-DMTL}, to tackle the hand image understanding tasks in the coarse-to-fine manner.
Our approach can efficiently exploit existing multi-modality datasets due to the MTL tactics and can harness enormous in-the-wild images through the self-supervised learning strategy, making it more practicable than previous methods.
The performance of our \textbf{HIU-DMTL} framework have been verified by extensive quantitative and qualitative experiments. 
Besides, most of the components are well explored and discussed in various ablation studies, which make it straightforward to explain why our framework can achieve good performance over HIU tasks. 
For further work, we hope the core designs of our HIU-DMTL framework can be adopted generally to other research domains. 

\newpage

{\small
\bibliographystyle{ieee_fullname}
\bibliography{Reference}
}

\end{document}